# Recurrence-Free Survival Prediction for Anal Squamous Cell Carcinoma Chemoradiotherapy using Planning CT-based Radiomics Model

Shanshan Tang*[1,2], Kai Wang*[1,2], David Hein[1], Gloria Lin[1], Nina N. Sanford[1], Jing Wang[1,2]

**Short title**: survival prediction with planning CT-based radiomics model

**Affiliations**

1 Department of Radiation Oncology, University of Texas Southwestern Medical Center, Dallas, TX

2 Medical Artificial Intelligence and Automation Laboratory, University of Texas Southwestern Medical Center, Dallas, TX

* co-first author

Corresponding author:

Jing Wang, Jing.Wang@UTSouthwestern.edu

Nina Sanford, Nina.Sanford@UTSouthwestern.edu

Address:

2280 Inwood Rd

Dallas, TX 75235

U.S.

Phone: 214-648-1795,

Fax:     214-645-8526

**Abstract**

**Objectives**: Approximately 30% of non-metastatic anal squamous cell carcinoma (ASCC) patients will experience recurrence after chemoradiotherapy (CRT), and currently available clinical variables are poor predictors of treatment response. We aimed to develop a model leveraging information extracted from radiation pretreatment planning CT to predict recurrence-free survival (RFS) in ASCC patients after CRT.

**Methods**: Radiomics features were extracted from planning CT images of 96 ASCC patients. Following pre-feature selection, the optimal feature set was selected via step-forward feature selection with a multivariate Cox proportional hazard model. The RFS prediction was generated from a radiomics-clinical combined model based on an optimal feature set with five repeats of five-fold cross validation. The risk stratification ability of the proposed model was evaluated with Kaplan-Meier analysis.

**Results**: Shape- and texture-based radiomics features significantly predicted RFS. Compared to a clinical-only model, radiomics-clinical combined model achieves better performance in the testing cohort with higher C-index (0.80 vs 0.73) and AUC (0.84 vs 0.79 for 1-year RFS, 0.84 vs 0.78 for 2-year RFS, and 0.86 vs 0.83 for 3-year RFS), leading to distinctive high- and low-risk of recurrence groups ($p<0.001$).

**Conclusions**: A treatment planning CT based radiomics and clinical combined model had improved prognostic performance in predicting RFS for ASCC patients treated with CRT as compared to a model using clinical features only.

**Introduction**

There has been a recent trend toward organ preservation in the treatment of an expanding list of gastrointestinal cancers, with anal squamous cell carcinoma (ASCC) serving as a historical model for this paradigm [1]. Since the publication of the Nigro trial in 1974 [2], definitive chemo-radiotherapy (CRT) has remained the standard of care for almost all patients with non-metastatic ASCC. While this strategy has successfully allowed most patients to avoid surgery, the treatment largely represents a one-size-fits all approach [3]. This leads to over and under-treatment for many patients resulting in suboptimal outcomes: patients who are over-treated experience excessive toxicity and those under-treated require surgical salvage with some eventually dying of their cancer. Fortunately, there are several ongoing efforts to personalize radiotherapy in ASCC: The PLATO trials tailor radiation therapy dose based upon clinical staging [4], the DECREASE trial randomizes patients with <4 cm and node negative ASCC to standard versus de-escalated chemoradiation (NCT04166318), and ECOG EA2165 assesses the addition of adjuvant nivolumab in locally advanced ASCC (NCT05002569). However, these studies all rely on initial TNM clinical stage for risk stratification which does not account for biological heterogeneity.

Radiomics [5], which involves transformation of imaging studies into mineable quantitative data, could improve cancer prognostication with studies modeling efficacy across multiple disease sites [6-10]. Studies have also demonstrated that genomics-based biomarkers are correlated with radiomic features, facilitating the evaluation of intratumoral genetic heterogeneity [11-14]. In ASCC, studies have assessed the utility of radiomics for prognostication based on MR [15-18] and PET/CT [19]. However, a baseline MRI is not standard of care in ASCC, and PET/CT are costly and occasionally delayed. In contrast, all patients receiving CRT undergo CT simulation for radiation planning, thus a CT-based radiomics approach could be highly feasible for predicting RFS in anal cancer. As such, the aim of this study was to evaluate whether radiomic features extracted from radiation planning CTs are predictive of treatment response permitting risk stratification in ASCC patients receiving CRT.

**Materials and Methods**

An RFS prediction model was constructed with radiomics features extracted from planning CT based on a Cox proportional hazard (CoxPH) model. Recurrence was defined as local or distant progression or recurrence after CRT. The overall workflow of the proposed method is summarized in Fig. 1.

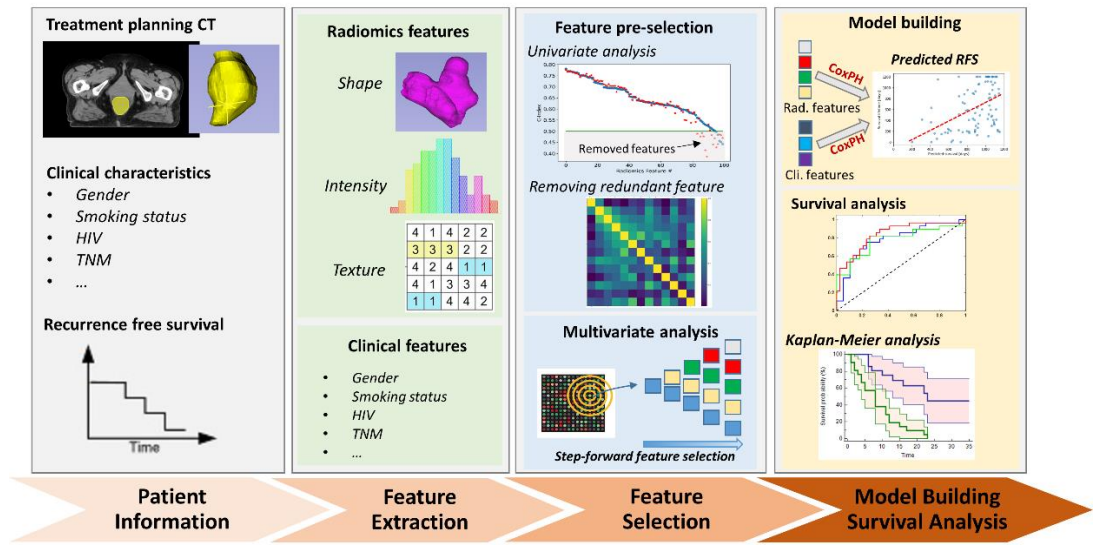

Figure 1. Overall workflow of recurrence-free survival (RFS) prediction using planning CT radiomics for patients with anal squamous cell carcinoma (ASCC). Rad., radiation; Cli., clinical. CoxPH, Cox proportional hazards model.

1. *Patient selection*

All patients treated with CRT for ASCC between 2010 and 2021 at UT Southwestern Medical Center were considered for the study. After patient exclusion (see Supplemental Fig. S1), 96 patients were included with a median follow up of 19.7 months (IQR: 7.8-35.3 months). Patients were treated with a median dose of 54 Gy (IQR: 52.2 to 55.4 Gy) over a median duration of 6 weeks (IQR: 5.4 to 6.5 weeks).

2. *Tumor delineation and image preprocessing*

Contours of primary gross tumor volume (GTVp) of the anal tumor were manually delineated on all axial slices from treatment planning CTs by the treating radiation oncologist as part of clinical care. In many patients, gas was noted to be included in the GTVp. For the purposes of this study, we excluded the low-intensity areas of gas within GTVp (HU cut-off: -150) before radiomics feature extraction (Supplemental Fig. S2). The processed tumor contours were checked by a radiation oncologist.

*3. Radiomics feature extraction and clinical feature collection*

Patients' gender, age, HIV status, pre-treatment CD4 count, smoking history, T stage, N stage and specific site of nodal metastases were recorded as clinical features. A detailed description of the patient cohort is provided in Supplementary Table S1. Of the 96 patients, there is 1 patient missing the N stage and specific site of nodal metastases values. The missing data in patient records were filled with the median value of the characteristics. Categorical clinical features were encoded to discrete values. Dates and dosage of CRT were recorded, as well as last date of follow up and date and site and date of recurrence, if applicable.

Planning CTs were performed on multiple CT scanners per department protocol (see details about imaging protocols in Supplemental Materials Table S2). A total of 100 radiomics features were extracted using an open-source radiomics toolbox, PyRadiomics [20], which meets the methodology and definitions of radiomics features in the Image Biomarker Standardization Initiative [21]. Extracted radiomics features includes shape and intensity indicators, as well as second-order textural features based on the gray level co-occurrence (GLCM), gray level run length (GLRLM), gray level size zone (GLSZM) and gray level dependence (GLDM) matrixes (See details in Supplemental Materials, section 4).

*4. Recurrence-free survival prediction*

We used the Lifelines Python package (Davidson-Pilon et al. 2019) to conduct survival analysis and to construct survival prediction models. Five-fold cross validation (5-CV) was performed for model construction and performance evaluation, with four folds used for training and the remaining fold for testing. Another 5-CV was applied inside the training group for feature selection, thus splitting it into 'training' and 'validation' sets in a 4:1 fashion. The predicted survival for each testing cohort in 5-CV was recorded and combined to form the prediction of the full patient cohort (Supplemental Fig. S2). Five repeats with random train-test split of 5-CV were done to minimize the influence of sample partition variations. The final prediction of RFS for each patient was the average of the predictions made from each repeat.

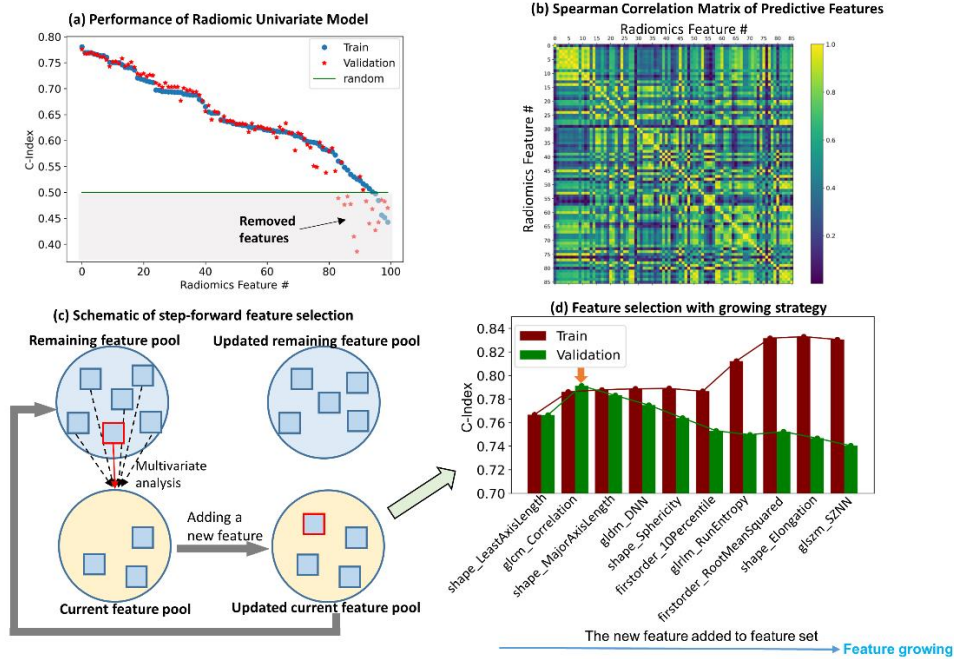

Fig. 2 Feature selection. (a) Performance of univariate model constructed for feature predictive power evaluation on training and validation cohorts, arranged as the descending order of C-index of the training cohort. (b) Spearman correlation matrix for feature similarity evaluation on the remaining features after exclusion of features with low predictive power evaluated in (a). (c) Illustration for step-forward feature selection. A new feature from the remaining feature pool that yields the best multivariate model performance was added to the current feature pool, so that the selected feature set was growing one feature each time. (d) Model performance with the step-forward feature selection strategy. The features on x-axis are the new features added to the current feature set. The final feature set was determined by the one that provided the best model performance in validation cohort, indicated by the orange arrow. DNN: dependence non-uniformity normalized. SZNN: size zone non-uniformity normalized.

4.1 Feature pre-selection

The predictive ability of each feature was evaluated by CoxPH model with C-index as the evaluation criterion. Features that generated C-index lower than a threshold (0.5 in this study) with either training or validation cohorts were excluded from the feature pool. As shown in Fig. 2(a), the predictive performance of radiomics features for training and validation cohorts were shown, arranged as the descending order of C-index of the training cohort. In the example of Fig. 2(a), of 100 radiomics features, 14 features were removed due to their low C-index with training and/or validation cohorts.

To remove redundant features, the intercorrelation of the remaining features were investigated using Spearman rank correlation analysis on training and validation data, as shown in Fig. 2(b). The cut-off for correlation-coefficient was empirically set to 0.8 to preserve robust features [22]. For a feature pair with a correlation coefficient higher than 0.8, the one with lower predictive ability, as determined in previous step, was excluded.

4.2 Step-forward feature selection and multivariate model construction

The optimal feature set for predicting RFS was determined via step-forward feature selection with multivariate CoxPH model. In each step, a new feature from the remaining feature pool was added to the current pool, a multivariate model was trained with training cohort, and model performance was validated with the validation cohort. A feature from the remaining feature pool that yields the best performance in the validation cohort was selected and added to the current feature pool. By doing this, the selected feature set grew one feature each time. Fig. 2(c) and 2(d) illustrate a schematic diagram of the step-forward feature selection as well as the multivariate model performance in training and validation cohorts for a given data split. The feature on x-axis of Fig. 2(d) is the new feature that was added to the current feature set. The final feature set was determined by the one that provided the best model performance in the validation cohort, as indicated by the orange arrow in Fig. 2(d). In this case, two radiomics features (shape least axis length and GLCM correlation) were selected to form a feature set for the radiomics-feature-based RFS prediction. The ceiling number of features in step-forward feature selection was limited to 10, as generally the number of features in a multivariate model should be less than 1/10 of the sample size [23]. A downward trend of C-index in validation cohort within several steps can be observed from Fig. 2(d), indicating the ceiling number of 10 features was sufficient to determine an optimal feature set.

4.3 Combination model for RFS prediction

With the determined feature set, the expected RFS time for each patient in the testing cohort was predicted and preserved. Predicted RFS for the full patient cohort were obtained after 5-CV and averaged after 5 repeats. The predicted RFS for every patient was calculated with the radiomics model and the clinical model separately, and the final RFS prediction for each patient was the average of the predicted RFS from the two models.

Because there are multiple ways of combining the radiomics and clinical models, other combination approaches were also evaluated: (1) we concatenated the radiomics features and clinical features all together for training and testing; (2) we pre-selected radiomics features and clinical features by removing low predictive and redundant features, then concatenated the pre-selected features together for multivariate analysis. The workflows of different ways to combine radiomics and clinical variables are included in Supplemental Fig. S6.

4.4 Model performance evaluation

The predicted RFS was compared to the observed outcome. The median predicted RFS of the full patient cohort was used as the cutoff to divide the patients into two groups. Kaplan-Meier analysis and log-rank test were then used to determine whether our model stratified patients into groups of significantly different RFS.

Binary assessments were performed on 1-year, 2-year, and 3-year RFS. Patients with follow-up time shorter than the given time-period were excluded for the binary assessment. The median RFS predictions of the according sub-patient cohorts were used as the cutoff for binary assessment.

**Results**

Of the 96 patients, 28 (29.2%) had disease progression after CRT. The number of patients included for 1-year, 2-year, and 3-year binary assessment were 85 (with 22 recurrences), 68 (with 28 recurrences), and 49 (with 28 recurrences), respectively.

Table I. Prediction performance comparison of radiomics-clinical combined model, clinical model, and radiomics model. Concordance index (C-index) was acquired with the full patient cohort, while binary assessment was made with 1-year, 2-year, and 3-year recurrence-free survival (RFS). CI: confident interval.

|  | C-index | 95% CI | AUC | Sensitivity | Specificity | Accuracy |
|---|---|---|---|---|---|---|
|  |  |  | *1-year RFS* | | | |
| **Combined Model** | 0.7953 | [0.67, 0.89] | 0.8391 | 0.9545 | 0.6508 | 0.7294 |
| **Clinical Model** | 0.7312 | [0.60, 0.85] | 0.7886 | 0.8636 | 0.619 | 0.6824 |
| **Radiomics Model** | 0.7545 | [0.62, 0.87] | 0.7706 | 0.8181 | 0.6032 | 0.6588 |
|  |  |  | *2-year RFS* | | | |
| **Combined Model** | 0.7953 | [0.67, 0.89] | 0.8397 | 0.8214 | 0.7179 | 0.7612 |
| **Clinical Model** | 0.7312 | [0.60, 0.85] | 0.7793 | 0.8214 | 0.7179 | 0.7612 |
| **Radiomics Model** | 0.7545 | [0.62, 0.87] | 0.7839 | 0.7500 | 0.6667 | 0.7015 |
|  |  |  | *3-year RFS* | | | |
| **Combined Model** | 0.7953 | [0.67, 0.89] | 0.8605 | 0.7857 | 0.8571 | 0.8163 |
| **Clinical Model** | 0.7312 | [0.60, 0.85] | 0.8299 | 0.7857 | 0.8571 | 0.8163 |
| **Radiomics Model** | 0.7545 | [0.62, 0.87] | 0.7619 | 0.7500 | 0.8095 | 0.7755 |

The combined radiomic-clinical model outperformed the clinical model in terms of C-index (0.80 vs 0.73) and AUC for all three evaluation time points (0.84 vs 0.79 for 1-year RFS, 0.84 vs 0.78 for 2-year RFS, and 0.86 vs 0.83 for 3-year RFS), as illustrated in Table I (definitions for the terms in Table I were described in Supplemental Materials). The C-index was acquired with the full patient cohort (n=96). The 95% confidence interval of the C-index for the three models were also evaluated. Although relatively wide ranges of 95% confidence intervals were observed due to small patient dataset, significant difference was found between the combined model and the clinical model (independent t-test, $p<0.001$) for their 95% confidence intervals of C-index. Binary assessments including calculations for AUC, sensitivity, specificity, and accuracy were based on the sub-patient cohorts with at least 1 year (n=85), 2 years (n=67), and 3 years (n=49) follow-ups.

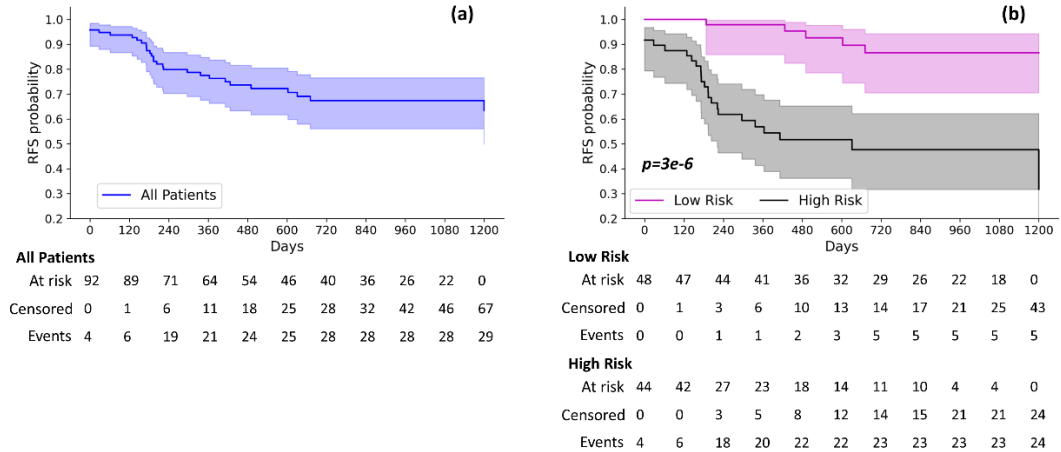

Fig. 3 Kaplan-Meier analysis of RFS curves. (a) RFS curve for the full patient cohort. (b) RFS curves for high- and low-risk groups identified with radiomics-clinical combined model. The median value of the predicted RFS was used to stratify the two risk groups, *p* value here was acquired via log-rank test.

Kaplan-Meier analysis and log-rank test were performed to illustrate the stratification for high- and low-risk groups, as shown in Fig. 3. Significant differences in survival probability were observed between the two groups.

For each experiment, there were 2 to 4 clinical features or 1 to 4 radiomics features selected in generating the clinical model or radiomics model, respectively, with training and validation cohorts. A list of selected features and their occurrences in the 5 times repeated 5-CV (a total of 25 experiments) is summarized in Table II. It is revealed that the T-stage in clinical model and least axis length in radiomics model were almost selected for every experiment, indicating their high and robust predictive ability. The predictive ability of a model can be further improved by adding one or several more features as a feature set. As illustrated by Fig. 2(d), the C-index in validation cohort improved from 0.76 to 0.79 when the radiomics model uses a feature set composed of least axial length and GLCM correlation.

Table II. Summary of selected features and their occurrences in the 25 experiments (5 repeats of 5-fold cross-validation).

| Feature set | feature name | Occurrence |
|---|---|---|
| Clinical | Primary Tumor Stage (T) | 25 |

|  | Lymph Node Stage (N) | 3 |
|---|---|---|
|  | CD4 Counts | 6 |
|  | Smoking Status | 12 |
|  | Inguinal Nodes | 2 |
|  | Age at Diagnosis | 7 |
|  | Gender | 1 |
| Radiomics | Shape Least Axis Length | 24 |
|  | Shape Flatness | 2 |
|  | GLCM Correlation | 3 |
|  | GLCM IDMN | 1 |
|  | First order 10th Percentile | 14 |
|  | First order Mean | 3 |
|  | First order Maximum | 1 |
|  | GLSZM Zone size entropy | 1 |

Figure 4 shows the ROC curves for RFS model evaluation. The ROC curve generated from the radiomics-clinical combined model was compared to ROC curves from the separate clinical and radiomics models in Fig. 4 for 1-year, 2-year, and 3-year RFS assessments. Although no significant improvement was achieved with the radiomics-clinical combined model compared to clinical model (Delong Test, $p$-values: 0.17 for 1-year RFS assessment, 0.097 for 2-year RFS assessment, and 0.43 for 3-year RFS assessment), a general trend was observed for all three time points showing that the model performance was improved with a higher AUC achieved by the combined model.

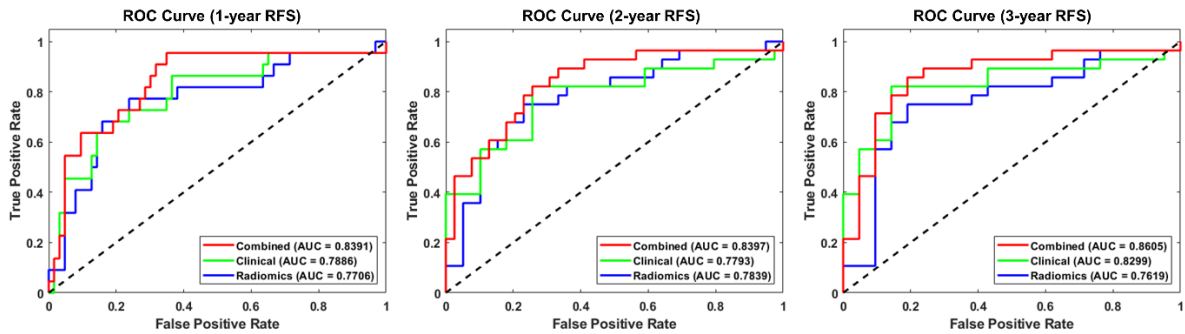

Fig. 4 ROC curves for model performance evaluation on 1-year, 2-year and 3-year recurrence classification. (a) Comparison of ROC curves and AUC among radiomics-clinical combined model, clinical model, and radiomics model

for 1-year RFS. (b) Comparison of ROC curves and AUC for 2-year RFS. (c) Comparison of ROC curves and AUC for 3-year RFS.

The radiomics model was also trained and tested using the original GTVp without the gas-exclusion step. Supplemental Table S3 and Fig. S3 compares the RFS prediction performances of using gas-excluded and original GTVp. Although no significant improvement was achieved with gas-excluded model from ROC curves ($p$-values: 0.64), a slightly higher C-index (0.7541 vs 0.7425) and 2-year AUC (0.7839 vs 0.7674) were observed with gas-excluded GTVp.

To remove redundant features, the cutoff for correlation coefficient was set to 0.8 in feature pre-selection step. Although this value was selected empirically, we noticed that the combined model outperformed the clinical model regardless of the cut-off selection in terms of C-index and AUC. For example, Supplemental Fig. S5 and Table S4 contain the model performances when the cut-off was set as 0.9. As pointed out by Van Timmeren et al., when assessing robustness of radiomics features with cut-off values of correlation coefficients, the number of "robust" features depend on the number of subjects involved [22]. Generally, a stricter/smaller cutoff of correlation coefficient leads to a feature set with less redundancy. In our study, the number of patients was modest (n=96). Therefore, a stricter cutoff was used to decrease the number of features for the further model generation.

In this study we obtained the final survival prediction by averaging the predicted RFS from clinical model and radiomics model. We also tried combining them by either concatenating all radiomics and clinical features together or concatenating pre-selected radiomics and clinical features, as described in the Methods section. C-index of 0.7512 and 0.7241 were received for the aforementioned two approaches. The proposed approach of using the averaged RFS predictions from radiomics and clinical predictions outperformed the aforementioned two approaches, as these two approaches lead to a higher probability of model overfitting as more features are included in model training with a limited data size. Detailed C-index and AUC for the different combination methods are listed in Table S5.

**Discussion**

In this paper, we assessed a novel RFS prediction method leveraging CT-based radiomics features in ASCC and found that our radiomics-clinical combined model achieved better performance as compared to a model with clinical variables only. The radiomics research quality score (RQS) [24] for this study is evaluated at 16 points (see Supplemental Table S7) which indicates the reliability of the study design and the reproducibility of the reported method, given that the median RQS value is 11.0 (IQR 3-12.75) [25].

Previously, prognostication in ASCC has relied on clinical characteristics alone. For instance, Wu et al. generated a nomogram model for survival using age, sex, ethnicity, marital status, and T and N stage, achieving a C-index of 0.68 [26]. Similarly, another model by Yang et al. includes age, sex, AJCC stage, and radiotherapy dose to construct a prognostic nomogram with C-index of 0.730 and AUC of 0.716 [27]. Clearly there is room for model improvement.

The concept underlying radiomics is that both morphological and functional clinical images contain qualitative and quantitative information, which may reflect the underlying pathophysiology of a tissue [24, 28]. Studies on hepatic metastases in colorectal cancer have found associations between CT-based radiomic features and KRAS gene mutations [11, 12]. Tumor textures extracted from CT images of non-small cell lung cancer showed a significant inverse association with tumor Glut-1 expression [13, 29]. However, relatively few studies have been published on RFS prediction of ASCC with radiomics, and all these studies were conducted with MR or PET/CT images. For example, MR-based radiomics features including skewness, Cluster Shade, baseline T2w Energy, etc., were found associated with 2-year RFS of ASCC after CTR [16-18]. Browns et al. found the GLCM entropy, NGLDM (Neighborhood Gray-Level Dependence Matrix) busyness and minimum CT value extracted from FDG-PET/CT in patient with ASCC were associated with CRT treatment outcome [19].

Notably, the radiomics predictors used to build our model were consistent with clinically used prognostic factors. For instance, the size of the tumor (least axis length as a radiomics feature) on CT image was found to be predictive. Additionally, our model discovered several new CT image-based features that are helpful in prediction of treatment outcome. For example, GLCM correlation, GLCM IDMN, and

GLSZM zone size entropy are markers of local homogeneity. Tumor heterogeneity reflects intra-tumoral architectural disorganization with many different intra-tumoral patterns such as area of necrosis, area of cell proliferation and area with important neo-angiogenesis, which provide insight into tumor biology [17].

Limitations of the study include restricting the imaging modality to CT images for feasibility; inclusion of other imaging modalities such as MRI and PET/CT could further improve the model's prediction performance. In addition, a minority of patients' scans were without IV contrast. This discrepancy in contrast use, could cause radiomic feature variation, although the comparison of median intensity in GTVp between contrast and non-contrast group shows no significant difference (Supplemental Fig. S7) [31].

**Conclusion**

In summary, we developed a radiomics-clinical combined RFS prediction model by incorporating treatment planning CT based radiomics features. Our result shows that the combined model outperformed the model with clinical features only, in terms of C-index (0.78 vs 0.73) and AUC evaluated at three time points (0.84 vs 0.79 for 1-year RFS, 0.84 vs 0.78 for 2-year RFS, and 0.86 vs 0.83 for 3-year RFS). Our model has the potential to assist in personalized management for local ASCC patients.

**Previous presentation of data**

Oral presentation at SWAAPM annual meeting (Galveston, 2023).

**Acknowledgement**

We acknowledge the funding support by National Institutes of Health (Grant No. R01CA251792).

# Supplementary Materials

**Title**: Recurrence-Free Survival Prediction for Anal Squamous Cell Carcinoma Chemo-radiotherapy using Planning CT based Radiomics Model

## 1. Patient inclusion

There were 228 patients treated with definitive chemo-radiotherapy (dCRT) for anal squamous cell carcinoma (ASCC) between 2010 and 2021 at UT Southwestern Medical Center. The year 2010 was chosen as the start date given initiation of intensity modulated radiation therapy (IMRT) at this time at our institution. Patients without pretreatment CT imaging stored in our system could not be included in this study. In addition, clinical exclusion criteria were multiple concurrent malignancies, prior pelvic radiation therapy, or inability to complete CRT to the planned definitive dose. All investigations in this study were carried out in accordance with the guidelines and regulations of institutional review board. The diagram of patient inclusion for this study is shown in Supplementary Fig. S1.

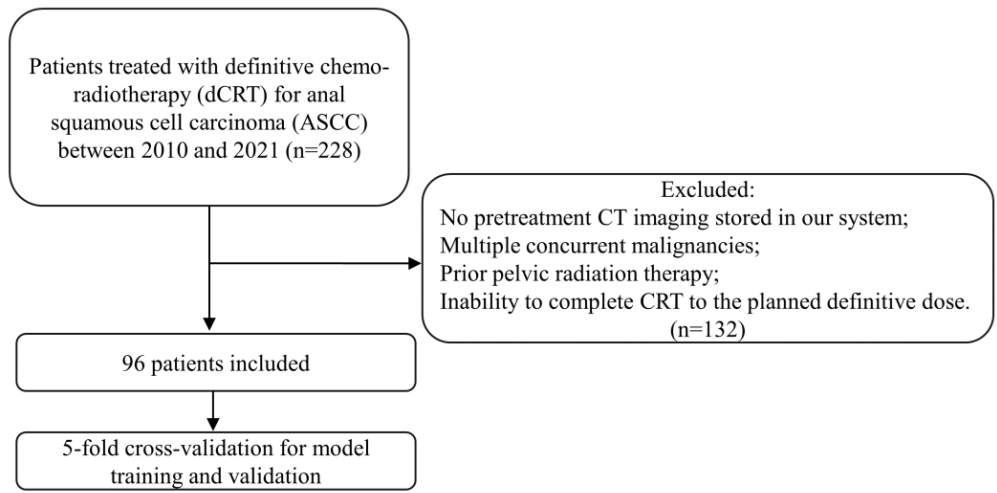

Figure S1: Diagram of the patient inclusion in this study.

## 2. Patient characteristics

Patients' gender, age, HIV status, pre-treatment CD4 count, smoking history, T stage, N stage and specific site of nodal metastases (inguinal, mesorectal, external and internal iliac) were recorded and used as features for clinical model. The detailed population description is provided in Table S1.

Table S1. Population descriptions for patient and tumor characteristics (n = 96)

| Characteristics | | Total | | | |
|---|---|---|---|---|---|
| | | *n* | *%* | *Median* | *std* |
| Gender | Male | 62 | 64.6 | | |
| | Female | 34 | 35.4 | | |
| Age at Dx [yrs] | | | | 54.5 | 12.1 |
| HIV | Yes | 45 | 46.9 | | |

| | No | 51 | 53.1 | | |
|---|---|---|---|---|---|
| CD4 count | | | | 1000 | 388.56 |
| Transplant recipient | Yes | 2 | 2.1 | | |
| | No | 94 | 97.9 | | |
| Smoking status | Current | 36 | 37.5 | | |
| | Former | 30 | 31.3 | | |
| | Never | 30 | 31.3 | | |
| Tumor stage | T1 | 15 | 15.6 | | |
| | T2 | 34 | 35.4 | | |
| | T3 | 32 | 33.3 | | |
| | T4 | 15 | 15.6 | | |
| Lymph node stage | N0 | 36 | 37.5 | | |
| | N1 | 34 | 35.4 | | |
| | N2 | 2 | 2.1 | | |
| | N3 | 23 | 23.9 | | |
| | Not Available | 1 | 1 | | |
| Metastatic disease stage | M0 | 93 | 96.9 | | |
| | M1 | 3 | 3.1 | | |
| Inguinal node involved | Yes | 47 | 49 | | |
| | No | 48 | 50 | | |
| | Not Available | 1 | 1 | | |
| Internal iliac involved | Yes | 20 | 20.8 | | |
| | No | 75 | 78.1 | | |
| | Not Available | 1 | 1 | | |
| external iliac involved | Yes | 24 | 25 | | |
| | No | 71 | 74 | | |
| | Not Available | 1 | 1 | | |
| Mesorectal involved | Yes | 28 | 29.2 | | |
| | No | 67 | 69.8 | | |
| | Not Available | 1 | 1 | | |

## 3. Imaging protocol

Each of the included patients in this study has a pre-treatment planning CT with a median interval of 11 days (IQR: 8-16 days) to the start day of their radiation therapy. As a single institution study, the imaging protocols are similar for all the patients. Most of the CT scans were acquired with Philips Brilliance Big Bore machine (86.46%), with pixel spacing of 1.1719 mm (72.92%), slice thickness of 2 mm (98.96%), tube voltage of 120 kVp (100%), tube current of 244 mA (65.63%), and exposure of 1230 mAs (70.83% ). We summarize the distribution of key parameters of imaging protocols for CT scan in Table S2 below.

Table S2: CT imaging protocols

| **Term** | **Number** | **Percentage (%)** |
|---|---|---|

| Manufacturer and Model | | |
|---|---|---|
| GE MEDICAL SYSTEMS HiSpeed NX/i | 1 | 1.04 |
| Philips Brilliance Big Bore | 83 | 86.46 |
| Mobius Imaging, LLC AIRO | 2 | 2.08 |
| Philips Big Bore | 10 | 10.42 |
| **Tube Voltage (kVp)** | | |
| 120 | 96 | 100.00 |
| **Current (mA)** | | |
| 100-200 | 3 | 3.13 |
| 200-250 | 74 | 77.08 |
| 300-400 | 8 | 8.33 |
| 400-500 | 7 | 6.25 |
| >500 | 4 | 4.17 |
| **Pixel Spacing (mm)** | | |
| IQR: 1.1719 – 1.3438 | 79 | 82.29 |
| **Thickness (mm)** | | |
| 3 | 95 | 98.96 |
| 1 | 1 | 1.04 |
| **Intravenous contrast agent** | | |
| With contrast | 61 | 63.5 |
| Without contrast | 35 | 36.5 |

### 4. Radiomic feature extraction

The radiomics features were extracted with PyRadiomic module (van Griethuysen, Fedorov et al. 2017)with Python version 3.9. All of the 100 radiomics features were in "original" category. There are six subcategories: Shape-based (14 features), First Order Statistics (18 features), Gray Level Co-occurrence Matrix (GLCM, 22 features), Gray Level Run Length Matrix (GLRLM, 16 features), Gray Level Size Zone Matrix (GLSZM, 16 features), and Gray Level Dependence Matrix (14 features).

### 5. Five-fold cross validation

A schematic illustration for five-fold cross validation (5-CV) can be seen in Fig. S2, where four folds were used for training and the remaining fold for testing. The predicted survival for each testing cohort in 5-CV was recorded and combined to form the prediction of the full patient cohort. This 5-CV repeated 5 times with random train-test split to minimize the influence of sample partition variations.

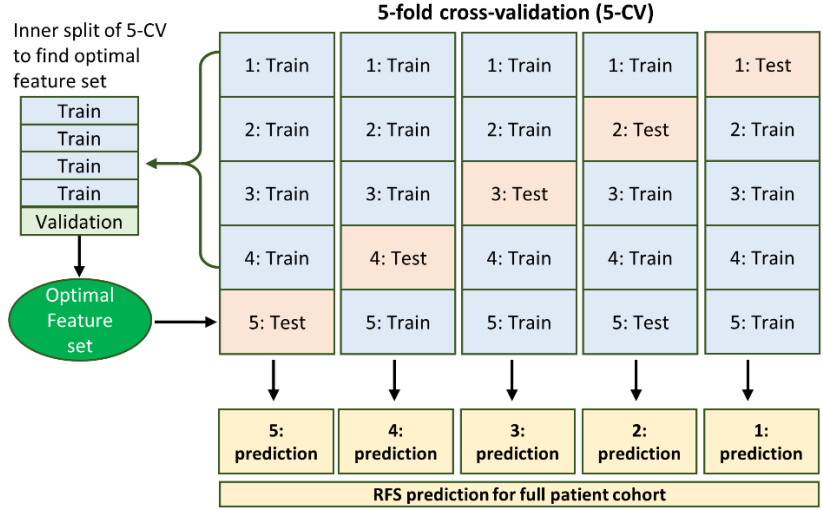

Figure S2. Illustration for five-fold cross validation (5-CV). The predicted survival for each testing cohort in 5-CV was recorded and combined to form the prediction of the full patient cohort.

## 6. Definitions of C-index, AUC, sensitivity, specificity, and accuracy

An ROC curve (receiver operating characteristic curve) is a graph showing the performance of a classification model at all classification thresholds. This curve plots two parameters: True Positive (TP) vs False Positive (FP). AUC stands for "Area under the ROC Curve." That is, AUC measures the entire two-dimensional area underneath the entire ROC curve. The definitions for sensitivity, specificity, and accuracy are as follows,

$$\text{sensitivity} = \frac{TP}{TP+FN} \quad (1)$$

$$\text{specificity} = \frac{TN}{TN+FP} \quad (2)$$

$$\text{accuracy} = \frac{TP+TN}{TP+FN+TN+FP} \quad (3)$$

, where FN and TN are the numbers of false negative and true negative.

## 6. Comparison of model performance between using GTVp with and without gas-exclusion

An example of GTVp delineation and gas-exclusion is shown in Fig. S3. Table S3 and Fig. S4 compares the RFS prediction performances of using gas-excluded and original GTVp. The general trend shows a slight performance improvement in terms of C-index and AUC with using gas-excluded GTVp.

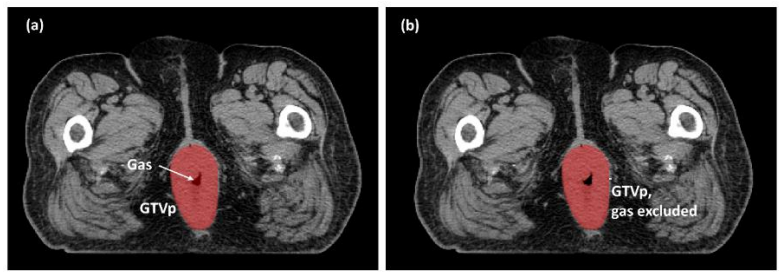

Figure S3. An example of gas-exclusion in primary gross tumor volume (GTVp). (a) Originally delineated GTVp with gas. (b) GTVp with gas excluded.

Table S3. Comparison of prediction performances of radiomics model with using original GTVp and gas-excluded GTVp.

|  | C-index | AUC | Sensitivity | Specificity | Accuracy |
|---|---|---|---|---|---|
| **GTVp w/ gas** | 0.7425 | 0.7674 | 0.7500 | 0.6667 | 0.7015 |
| **GTVp w/o gas** | 0.7541 | 0.7839 | 0.7500 | 0.6667 | 0.7015 |

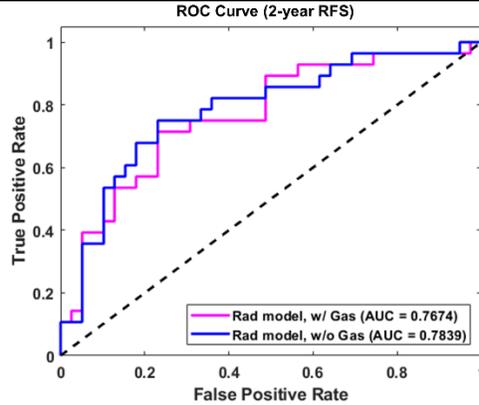

Figure S4. ROC curves for radiomics model performance comparison between using GTVp with and without gas exclusion. (a) ROC curve generated using GTVp that contains gas. (b) ROC curve generated using gas-excluded GTVp. AUC, area under the curve.

## 7. Model performance with different cutoff for feature pre-selection

In feature pre-selection, the intercorrelation of features were investigated using a Spearman rank correlation matrix. A cutoff of correlation-coefficient was determined empirically to remove redundant features. This cutoff was selected to be slightly strict as 0.8 in this study due to the limited patient size. We also tested with a cutoff of 0.9 and summarized the model performance in Table S4 and Fig. S5. It was observed that the different cutoff only affected the radiomics model but not clinical model, as the clinical characteristics are much more independent. With a less strict cutoff, the prediction performance still outperforms the clinical model in terms of C-index and AUC.

Table S4. Comparison of different cutoff for intercorrelation-coefficient of features in feature pre-selection.

|  | pre-selection cutoff 0.8 | | pre-selection cutoff 0.9 | |
|---|---|---|---|---|
|  | C-index | AUC | C-index | AUC |
| **Combined Model** | 0.7953 | 0.8397 | 0.7977 | 0.8342 |
| **Clinical Model** | 0.7312 | 0.7793 | 0.7312 | 0.7793 |
| **Radiomic Model** | 0.7545 | 0.7839 | 0.735 | 0.7582 |

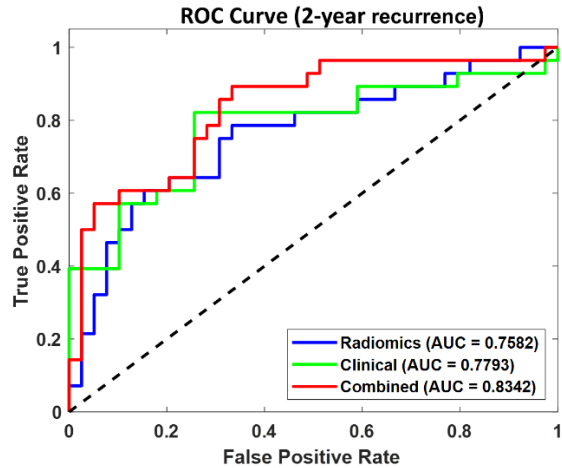

Figure S5. Comparison among combined model, clinical model, and radiomics model when a less strict cut-off (0.9) in intercorrelation coefficient used for feature selection. Combined model still outperformed clinical model in terms of AUC (as well as C-index).

## 8. Combined model generation from radiomics and clinical features

There are different ways to combine the radiomics features and clinical features into a combined model for RFS prediction. The workflows were summarized in Fig. S6. Figure S6(a) shows the approach used for this study, where RFS prediction generated from radiomics and clinical features separately and the final RFS prediction is an average of the predictions from radiomics and clinical features. The other two approaches tested in this study are, radiomics and clinical features combined together at the beginning for model construction (Fig. S6(b)) and radiomics and clinical features pre-selected and then combined for model construction and RFS prediction (Fig. S6(c)). RFS prediction performance of the radiomics-clinical combined models constructed with different methods were summarized in Table S5.

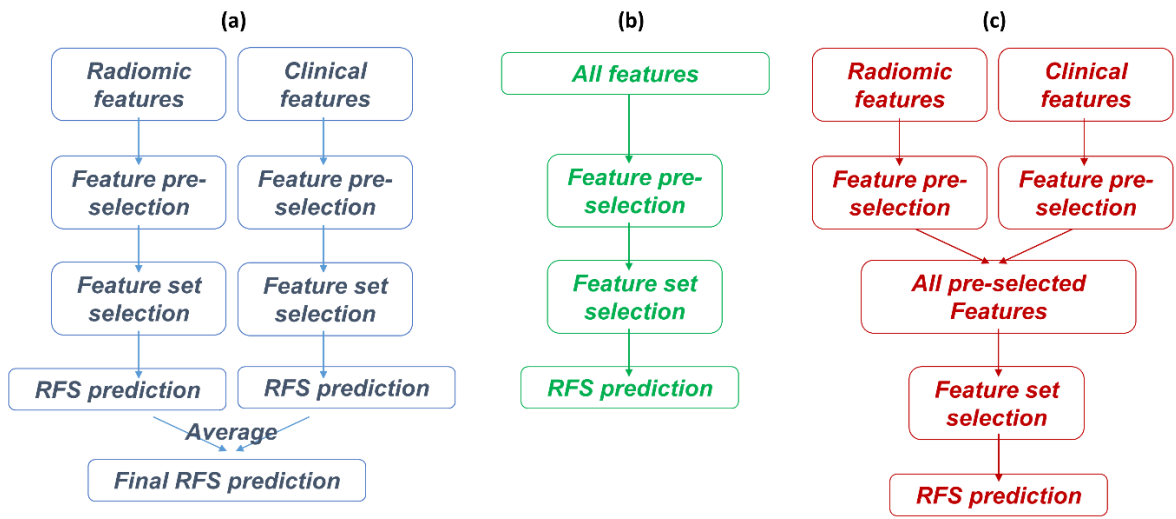

Figure S6. Combined model generation and recurrence free survival (RFS) prediction. (a) RFS prediction generated from radiomics feature and clinical features separately. The final RFS prediction is an average of the predictions from radiomics and clinical features. (b) Radiomics and clinical features were combined together at the beginning

for model construction. (c) Radiomics and clinical features were pre-selected and combined for model construction and RFS prediction.

Table S5. RFS prediction performance of the radiomics-clinical combined models constructed with different methods. Approach (a) was constructed by averaging the predictions from radiomics model and clinical models. Approach (b) was constructed by combining all the radiomics and clinical features at the beginning. Approach (c) was combining the pre-selected radiomics and clinical features for model construction.

|  | Combined Model | | |
| --- | --- | --- | --- |
|  | Approach (a) | Approach (b) | Approach (c) |
| **C-Index** | 0.7953 | 0.7512 | 0.7241 |
| **AUC** | 0.8397 | 0.7607 | 0.7329 |

## 9. Comparison between contrast and non-contrast groups

The median intensity inside the primary gross tumor volume (GTVp) was extracted and compared for patients with and without intravenous contrast agent administrated before CT scanning, as shown in Fig. S7. There were no significant difference was observed from the collected data.

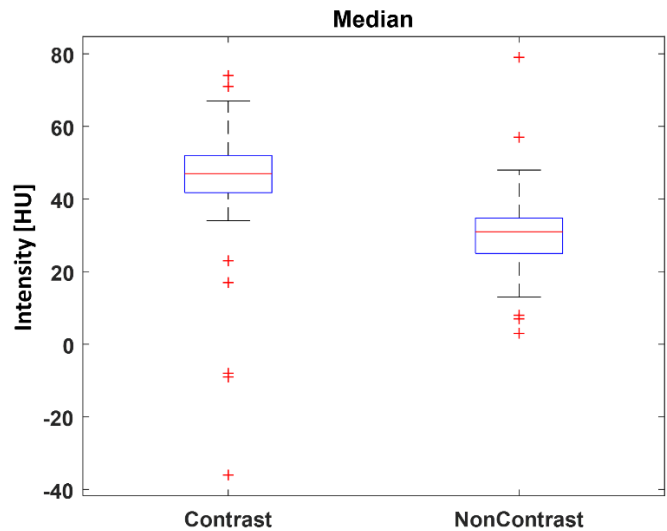

Figure S7. Median value comparison for patient with and without administration of contrast agent.

## 10. Checklist for radiomics research quality score

We reviewed our study according to the checklist for radiomics research quality score (QRS, see Table S6) (Lambin, Leijenaar et al. 2017), which is a score that assesses the characteristics and ultimately the quality of a radiomics study including the reporting of it. The section/topic we covered in the study is scored according to the description. Our QRS was evaluated as 16 points, while the reported median RQS score by Park et al. was 11.0 (IQR 3-12.75) (Park, Kim et al. 2020).

Table S6. Checklist for radiomics research quality score

| Criteria | Description | Points | Our Points |
| --- | --- | --- | --- |
| 1 | Image protocol quality - well-documented image protocols (for example, contrast, slice thickness, energy, etc.) and/or usage of public image protocols allow reproducibility/replicability | + 1 (if protocols are well-documented) + 1 (if public protocol is used) | 1 |
| 2 | Multiple segmentations - possible actions are: segmentation by different physicians/algorithms/software, perturbing segmentations by (random) noise, segmentation at different breathing cycles. Analyse feature robustness to segmentation variabilities | +1 | 0 |
| 3 | Phantom study on all scanners - detect inter-scanner differences and vendor-dependent features. Analyse feature robustness to these sources of variability | +1 | 0 |
| 4 | Imaging at multiple time points - collect images of individuals at additional time points. Analyse feature robustness to temporal variabilities (for example, organ movement, organ expansion/shrinkage) | +1 | 0 |
| 5 | Feature reduction or adjustment for multiple testing - decreases the risk of overfitting. Overfitting is inevitable if the number of features exceeds the number of samples. Consider feature robustness when selecting features | - 3 (if neither measure is implemented) + 3 (if either measure is implemented) | 3 |
| 6 | Multivariable analysis with non radiomics features (for example, EGFR mutation) - is expected to provide a more holistic model. Permits correlating/inferencing between radiomics and non radiomics features | +1 | 1 |
| 7 | Detect and discuss biological correlates - demonstration of phenotypic differences (possibly associated with underlying gene–protein expression patterns) deepens understanding of radiomics and biology | +1 | 0 |
| 8 | Cut-off analyses - determine risk groups by either the median, a previously published cut-off or report a continuous risk variable. Reduces the risk of reporting overly optimistic results | +1 | 1 |
| 9 | Discrimination statistics - report discrimination statistics (for example, C-statistic, ROC curve, AUC) and their statistical significance (for example, p-values, confidence intervals). One can also apply resampling method (for example, bootstrapping, cross-validation) | + 1 (if a discrimination statistic and its statistical significance are reported) + 1 (if a resampling method technique is also applied) | 2 |
| 10 | Calibration statistics - report calibration statistics (for example, Calibration-in-the-large/slope, calibration plots) and their statistical significance (for example, P-values, confidence intervals). One can also apply resampling method (for example, bootstrapping, cross-validation) | + 1 (if a calibration statistic and its statistical significance are reported) + 1 (if a resampling method technique is also applied) | 0 |
| 11 | Prospective study registered in a trial database - provides the highest level of evidence supporting the clinical validity and usefulness of the radiomics biomarker | + 7 (for prospective validation of a radiomics signature in an appropriate trial) | 0 |
| 12 | Validation - the validation is performed without retraining and without adaptation of the cut-off value, provides crucial information with regard to credible clinical performance | - 5 (if validation is missing) + 2 (if validation is based on a dataset from the same institute) + 3 (if validation is based on a dataset from another institute) + 4 (if validation is based on two datasets from two distinct institutes) + 4 (if the study validates a previously published signature) + 5 (if validation is based on three or more datasets from distinct institutes) *Datasets should be of comparable size and should have at least 10 events per model feature | 2 |
| 13 | Comparison to 'gold standard' - assess the extent to which the model agrees with/is superior to the current 'gold standard' method (for example, TNM-staging for survival prediction). This comparison shows the added value of radiomics | +2 | 2 |
| 14 | Potential clinical utility - report on the current and potential application of the model in a clinical setting (for example, decision curve analysis). | +2 | 2 |
| 15 | Cost-effectiveness analysis - report on the cost effectiveness of the clinical application (for example, QALYs generated) | +1 | 1 |
| 16 | Open science and data - make code and data publicly available. Open science facilitates knowledge transfer and reproducibility of the study | + 1 (if scans are open source) + 1 (if region of interest segmentations are open source) + 1 (if code is open source) + 1 (if radiomics features are calculated on a set of representative ROIs and the calculated features and representative ROIs are open source) | 1 |